\documentclass[sigconf]{acmart}

\AtBeginDocument{%
  \providecommand\BibTeX{{%
    \normalfont B\kern-0.5em{\scshape i\kern-0.25em b}\kern-0.8em\TeX}}}
    
\copyrightyear{2019}
\acmYear{2019}
\setcopyright{acmlicensed}
\acmConference[FDG '19]{FDG '19: Procedural Content Generation Workshop}{August 26-30, 2019}{San Luis Obispo, CA}
\acmBooktitle{FDG '19: Procedural Conten Generation Workshop, August 26-30, 2019, San Luis Obispo, CA}
\acmPrice{15.00}
\acmDOI{10.1145/1122445.1122456}
\acmISBN{978-1-4503-9999-9/18/06}

\usepackage{graphicx}
\usepackage{subcaption}

\begin{document}

\title{Two-step Constructive Approaches for Dungeon Generation}

\author{Michael Cerny Green}
\affiliation{%
  \institution{Game Innovation Lab\\New York University}
  \streetaddress{5 Metrotech Center}
  \city{Brooklyn} 
  \state{NY} 
  \postcode{11201}
}
\email{mcg520@nyu.edu}

\author{Ahmed Khalifa}
\affiliation{%
  \institution{Game Innovation Lab\\New York University}
  \streetaddress{5 Metrotech Center}
  \city{Brooklyn} 
  \state{NY} 
  \postcode{11201}
}
\email{ahmed@akhalifa.com}

\author{Athoug Alsoughayer}
\affiliation{%
  \institution{Game Innovation Lab\\New York University}
  \streetaddress{5 Metrotech Center}
  \city{Brooklyn} 
  \state{NY} 
  \postcode{11201}
}

\author{Divyesh Surana}
\affiliation{%
  \institution{Game Innovation Lab\\New York University}
  \streetaddress{5 Metrotech Center}
  \city{Brooklyn} 
  \state{NY} 
  \postcode{11201}
}

\author{Antonios Liapis}
\affiliation{%
  \institution{Institute of Digital Games\\University of Malta}
  \city{Msida}
  \country{Malta}
}
\email{antonios.liapis@um.edu.mt}

\author{Julian Togelius}
\affiliation{%
  \institution{Game Innovation Lab\\New York University}
  \streetaddress{5 Metrotech Center}
  \city{Brooklyn} 
  \state{NY} 
  \postcode{11201}
}
\email{julian@togelius.com}

\renewcommand{\shortauthors}{Green et al.}

\begin{abstract}
This paper presents a two-step generative approach for creating dungeons in the rogue-like puzzle game MiniDungeons 2. Generation is split into two steps, initially producing the architectural layout of the level as its walls and floor tiles, and then furnishing it with game objects representing the player's start and goal position, challenges and rewards. Three layout creators and three furnishers are introduced in this paper, which can be combined in different ways in the two-step generative process for producing diverse dungeons levels. Layout creators generate the floors and walls of a level, while furnishers populate it with monsters, traps, and treasures. We test the generated levels on several expressivity measures, and in simulations with procedural persona agents.
\end{abstract}

%
%



\maketitle

\section{Introduction}\label{sec:introduction}
While research on level generation focuses on level generators based on stochastic search~\cite{togelius2011search}, constraint solving~\cite{smith2011answer,smith2011tanagra}, or machine learning~\cite{summerville2017procedural}, level generation in published games is mostly carried out via constructive algorithms. Unlike generate-and-test processes, constructive generators do not evaluate and re-generate output; for example, cellular automata and grammars can be used for constructive generation, as well as more freeform approaches such as diggers~\cite{shaker2016constructive}. Such generators are computationally lightweight since they do not evaluate their generated output. This allows games to quickly create endless variations to game-play by generating maps as in \emph{Minecraft} (Mojang 2011), weapons as in \emph{Borderlands} (Gearbox 2009) or NPCs as in \emph{Skyrim} (Bethesda 2011) in real-time. However, choosing the right algorithm for the design constraints seems to be an art rather than a science. A better understanding of the properties of different families of generators as well as how they can be combined could help advance this situation.

One way to better understand the properties of constructive generation techniques is to systematically investigate their differing performance when applied to different aspects of level generation. For example, we can divide up the task of generating a level into different phases and use different constructive algorithms for each phase. Multi-agent generation processes have been explored for terrain generation \cite{doran2010controlled}; however, in this case, generation was controlled by multiple agents given explicit areas/types of terrain to generate, and agents could manipulate each others' finished products. In this paper, the second step of the process builds off of---but does not attempt to change---the result of the first.

Many mobile games use procedural content generation (PCG) to quickly generate content---with varying degrees of success. Flappy Bird (dotGears 2013), Doodle Jump (Lima Sky 2009), Downwell (Moppin 2015), Polytopia (Midjiwan AB 2016), and Temple Run (Imangi Studios 2011) are just a few of many examples. This paper compares several constructive generation approaches that produce levels for the rogue-like puzzle game MiniDungeons 2~\cite{holmgard2015minidungeons2}.
The fast generation afforded by constructive approaches allows the game to create new levels with minimal lag even on a mobile device, for which MiniDungeons 2 is intended. The novelty of this approach is the use and analysis of a two-step process for generating the level's architecture first, using a standalone \emph{layout creator}, and distributing the game objects on that architecture based on game-specific rules using a \emph{furnisher}. This allows for different combinations of creators and furnishers and can also work with manually created architectures (furnished automatically) or vice versa. While creators only place walls or floor tiles and use tried-and-tested algorithms popular in rogue-like dungeon generation~\cite{shaker2016constructive}, furnishers must account for the interactions and dynamics of different game objects. Moreover, since constructive generators do not test the final result, the rules used in the different furnishers must ensure that the level can be completed but also viable for different playstyles. To test the latter, artificial agents acting as play personas from prior work~\cite{holmgard2018automated} are used to test the generated levels. The variety of personas allows us to assess the generated content from multiple perspectives.

The contributions of this paper include the division of dungeon generation into two phases and several algorithms that work in each phase; a novel agent-based furnisher; and play-testing the resulting levels with procedural personas.

\section{Background}\label{sec:relatedwork}
We start our exploration in procedural level generation of Minidungeons 2 by reviewing how constructive map generation has been done in other games. Based on Shaker et al.~\cite{shaker2016constructive}, popular methods for generating dungeons include binary state partitioning, cellular automata and digger agents. 
A \textit{digger agent} is placed in a dungeon filled with impassable blocks (often in a random position), and removes the block it is in while moving to adjacent tiles following random or rule-based strategies. 
\textit{Cellular automata} are popular methods used to generate organic and smooth-looking maps, including islands and caves~\cite{johnson2010cellular}. Cellular automata work in iterative steps; in each step they change a tile based on patterns in its adjacent tiles. Rules regarding adjacent tiles and how they affect the current tiles can be elaborate or include stochasticity; the most straightforward way to create caves, however, is to change the current tile to match the majority of its adjacent tiles~\cite{shaker2016constructive}.

\textit{Grammars} have proven to be successful in generating adventure game levels due to their formal structure which can be intuitively interpreted and edited by human designers. Dormans~\cite{dormans2011level} uses graphs to generate missions as an initial description of a level, then transforming that into the rendered space. The whole process can result in a highly automated yet controlled way for map generation. This research was extended with a highly controllable automated system of model transformations \cite{dormans2011level}.

While constructive methods take a straightforward approach to generation and have a long history in the game industry, to the best of our knowledge no one has extensively analyzed the combination of different methods that iteratively generate map layouts and distribute game objects. This paper presents nine unique creator-furnisher combinations and analyzes patterns in resulting levels.

\section{The MiniDungeons 2 Game}\label{sec:game}
Minidungeons 2 (MD2) is a 2D rogue-like dungeon crawler~\cite{holmgard2015minidungeons2} in which the player controls a hero and tries to find the exit of a dungeon, while encountering a variety of monsters and objects along the way. The level is set on a $10{\times}20$ tile grid, where each tile is either impassible (wall) or passable (floor). Floor tiles can contain interactable objects (\textit{treasures, potions, traps, portals}) or game characters, subdivided into enemies (\textit{goblins, goblin mages, blobs, ogres, minitaurs}) and the \textit{hero} controlled by the player. LOS stands for Line of Sight, i.e. if this entity has a clear sight-line to the designated target.

\begin{table}
\small
\begin{tabular}{| p{5.5em} | p{21.5em} |}
\hline
Goblin & Move towards hero if in LOS \\\hline
Goblin Mage & Hurl bolts at hero in LOS if within 3 tiles\\\hline
Blob & Move toward closest hero or potion if in LOS; power up if colliding with potion or another blob (removing collided object)\\\hline
Ogre & Move towards closest hero or treasure if in LOS; empty treasure if colliding with it\\\hline
Minitaur & Move towards hero anywhere in the dungeon, following A* pathfinding; can not be killed, only stunned for 3 rounds\\
\hline
\end{tabular}
\caption{Movement strategies of monsters.}
\label{table:monster_behavior}
\end{table}

All game characters have a preset number of hit-points (HP) and deal damage when they collide with the hero or each other; goblin mages also hurl bolts that deal damage at range. When a game character runs out of HP, they die. To win, the player must move the hero to the exit without dying; the hero moves to adjacent tiles, except if stepping into a portal in which case it instantly transports to the other portal (a level has two portals, or none). Unlike its predecessor~\cite{holmgard2014evolvingpersonas}, in MD2 monsters' damage and movement behaviors are deterministic (see Table \ref{table:monster_behavior}). This brings MD2 closer to a puzzle game where the player must plan ahead how to reach the exit without dying as well as collecting as much treasure or killing as many monsters as possible. The personas used in the analysis of the creator-furnisher combinations are done using agents from another project in Minidungeons 2~\cite{holmgard2018automated}.

\section{Constructive Generators for MD2}\label{sec:generators}
The generative process for MD2 levels is split into two steps: the first step generates the architectural layout of the dungeon; the second step furnishes it with game objects and monsters. Three generators are built for the first step, identified as \emph{layout creators}, and determine which tiles in the dungeon will be passable (floor) or impassable (wall). Once the layout has been created (in whichever fashion), it is furnished with game objects (monsters, treasure, potions, traps, portals, the entrance and the exit). Three furnishers are introduced and evaluated in this paper, identified as \emph{Game Element furnishers}.

\subsection{Layout Creators}\label{sec:generators_architecture}
Based on Shaker et al.~\cite{shaker2016constructive}, three layout creators were designed following popular methods for constructive approaches: constraint-based, cellular automata, and agent-based creators. All creators ignore the border of the $10{\times}20$ MD2 grid which is always filled with walls, and operate on a grid of $8{\times}18$ tiles.

\subsubsection{Constraint-based:}
The constraint-based creator (CC) is inspired by the TinyKeep dungeon generator and collision detection systems~\cite{adonaac2015dungeon}. The generator spawns a random number of rooms with a random width and height. At every step, the locations of the rooms are modified by separating colliding rooms from each other, either until no more collisions occur and all the rooms are within the bounds of the map or until 100 iterations have passed. This generator initializes anywhere from 8 to 16 rooms with a width between 4 to 6 tiles and a height between 4 to 8 tiles.

\subsubsection{Cellular Automata:}
Based on the binary cellular automata cave generator in Shaker et al.~\cite{shaker2016constructive}, the MD2 cellular automata creator (CAC) initially populates all $8{\times}18$ tiles with either wall (45\% chance) or floor tiles. Cellular automata changes each tile's type to the type of the majority of its neighbors (considering the 8 closest neighboring tiles). In MD2, this resulted in a single `island' of floor tiles in the center. To counter this, a rule was added that checks if the map consists of more than 75\% floor tiles, in which case the map is filled with more wall tiles and another step of cellular automata is applied. This process will continue until there are less than 75\% floor tiles. This method can create multiple isolated `islands' of free space, which the player would be unable to reach. The largest empty space takes precedence. Any smaller islands (defined as isolated, empty tiles smaller than the largest space) are filled with walls.

\subsubsection{Agent-based:}
The agent-based creator (AC) was formulated much like the digger agent described by Shaker et al.~\cite{shaker2016constructive}. The $8{\times}18$ tile grid initially is filled with walls. The agent is then placed on a random tile, converting it to a floor tile. The agent then randomly selects a direction to travel in, moves forward one tile and converts it into a floor tile. The process continues in steps; in every step the agent has a chance to change direction, increasing the probability ($+5\%$) each step that it does not. This process continues until the map contains a number of floor tiles, which is randomly selected in the beginning to be within the range of 75 to 95 tiles.

\subsection{Game Element Furnishers}\label{sec:generators_gameelements}
Once the architecture is created by one of the above generators (or by a human creator), all game elements are added through another process identified as the \emph{furnisher}. Three furnishers are used in this paper; their differences are primarily in the rules for placing objects, and whether objects can change their location after being placed. Game elements added by the furnishers are the entrance (where the hero starts from), exit, treasures, potions, portals, traps and monsters.

\subsubsection{Constraint-based:}

\begin{table}
\small
\begin{tabular}{| p{5.5em} | p{21.5em} |}
\hline
Entrance & Always within 8 tiles of one end of the \textbf{LP}\\\hline
Exit & Within 5 tiles of the opposite end of the \textbf{LP} from the entrance\\\hline
Treasure Chests & Surrounded by 2 or more walls, with a preference towards 3 walls\\\hline
Potions & Scattered randomly across the map\\\hline
Portals & One placed 5-10 tiles from Entrance, the other 5-10 tiles from Exit; at least 10 tiles from each other\\\hline
Traps & On or around (within 1 tile) the shortest path between the Entrance and Exit\\\hline
Goblins & One side be a wall\\\hline
Goblin~Mages & Must be adjacent to a Goblin\\\hline
Ogres & 4-8 tiles away from a Treasure chest in LOS\\\hline
Blobs & 4-8 tiles away from a Potion in LOS\\\hline
Minitaur & 4-8 tiles away from Entrance\\
\hline
\end{tabular}
\caption{Constraint-based furnisher object placement rules}\label{table:constraint-based-furnisher-heuristics}
\end{table}

In the constraint-based furnisher (CF), each game element is constrained to areas of the map with specific characteristics. Before placing any object, the furnisher finds the \textit{longest path} (\textbf{LP}) that exists between any two points on the map. Elements are added iteratively, with specific elements such as the entrance and exit added first. As each element is added, the map is scanned for suitable locations that satisfy that element's constraints. The furnisher then randomly selects a suitable location among these to place the element. In-depth rules are found in Table~\ref{table:constraint-based-furnisher-heuristics}. 

\subsubsection{Cellular Automata:}

\begin{table}
\small
\begin{tabular}{| p{5.5em} | p{21.5em} |}
\hline
Entrance/Exit & Any tile where the other object is not present in a 5 tile neighborhood\\\hline
Treasure Chests & At least 3 neighbors must be walls using 1 tile neighborhood\\\hline
Potions & At most 3 neighbors are populated with objects using 1 tile neighborhood\\\hline
Portals & One portal must neighbor the Entrance, the other portal must neighbor the Exit using 3 tile neighborhood\\\hline
Traps & 5 or more neighbors must be populated with walls or other objects using 1 tile neighborhood\\\hline
Goblins & 4 or more neighbors are walls using 1 tile neighborhood, and none of the tiles in a 3 tile neighborhood are goblins\\\hline
Goblin~Mages & 1 or more neighbors must be a Goblin using 3 tiles neighborhood\\\hline
Ogres & None of the tiles in 1 tile neighborhood can be walls\\\hline
Blobs & At least 1 of the tiles in a 3 tile neighborhood is a Potion\\\hline
Minitaur & 1 of the tiles in a 3 tile neighborhood is the Entrance\\
\hline
\end{tabular}
\caption{Cellular Automata furnisher object placement rules}\label{table:cellular-automata-furnisher-heuristics}
\end{table}

Similar to the CA creator, the cellular automata furnisher (CAF) populates a map with game objects based on each tile's state and its neighboring tiles' states. The difference between the furnisher and the creator is that the furnisher visits tiles in random order instead of sequential, has a variable size neighborhood which allows CAF to access tiles that are more than 1 tile away, and has a restriction on the number of each placed object (one entrance, two traps, etc) so it will not overpopulate the dungeon. All floor tiles in the map are checked iteratively; if the tile is empty and fulfills the neighborhood requirements of Table~\ref{table:cellular-automata-furnisher-heuristics}, the corresponding game object is added to map. Since the neighborhood requirements are mutually exclusive, a tile can have only one object type.

\subsubsection{Agent-based:}

\begin{table}
\small
\begin{tabular}{| p{5.5em} | p{21.5em} |}
\hline
Entrance/Exit & Move as far apart from each other as possible\\\hline
Treasure Chests & Moves closer to goblins in Line-Of-Sight (LOS)\\\hline
Potions & Move randomly around the map\\\hline
Portals & Move as far apart from each other, the Entrance, and the Exit\\\hline
Traps & Move away from other traps and goblins in LOS. Move towards treasure in LOS\\\hline
Goblins & Move away from other goblins in LOS\\\hline
Goblin~Mages & Move toward goblins in LOS and away from other Goblin-Mages within LOS\\\hline
Ogres & Move away from Ogres in LOS (within 6 tiles), and move within 4 tiles of Treasure in LOS\\\hline
Blobs & Move within 4 tiles of other Blobs and Potions in LOS\\\hline
Minitaur & Move as far away as possible from the Entrance and Exit\\
\hline
\end{tabular}
\caption{Agent-Based furnisher object placement rules}\label{table:agent-furnisher-heuristics}
\end{table}

The Agent-based furnisher (AF) differs from previous ones as each object takes turns moving around the map by itself. Every game element is given its own heuristic for movement priorities (see Table~\ref{table:agent-furnisher-heuristics}) and is randomly placed somewhere on the map. Every ``turn'', all objects move around the map according to what they see and their proximity to other objects. For example, treasures actively seek out Goblins to guard them, and Goblins attempt to hide from other Goblins. All objects operate on a simple one-step-lookahead.  After 45 turns of movement (based on preliminary testing), the map is considered furnished.

\section{Evaluation}\label{sec:evaluation}
In order to evaluate the different patterns favored by each of the three creators and the three furnishers, each creator produces 1000 layouts for each furnisher, which then furnishes each layout once.
The result is 9000 MD2 levels for all creator-furnisher combinations. Statistical tests in this section use $p<0.05$ via Student's two-tailed $t$-test assuming unequal variance.

\subsection{Differences between generators}

\begin{figure}
    \centering
    \begin{subfigure}[t]{0.3\linewidth}
        \centering
        \includegraphics[width=.9\textwidth]{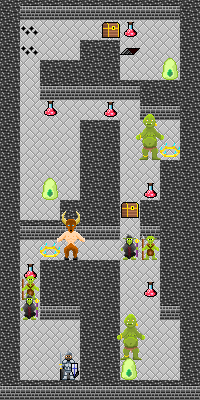}
        \caption{CC-CF}
    \end{subfigure}
    \begin{subfigure}[t]{0.3\linewidth}
        \centering
        \includegraphics[width=.9\textwidth]{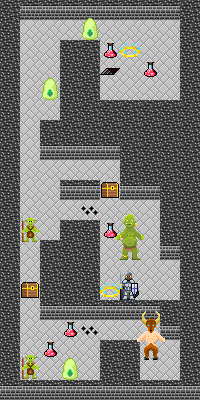}
        \caption{CC-CAF}
    \end{subfigure}
    \begin{subfigure}[t]{0.3\linewidth}
        \centering
        \includegraphics[width=.9\textwidth]{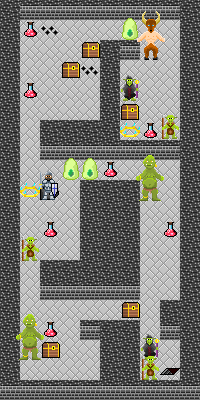}
        \caption{CC-AF}
    \end{subfigure}
    
    \begin{subfigure}[t]{0.3\linewidth}
        \centering
        \includegraphics[width=.9\textwidth]{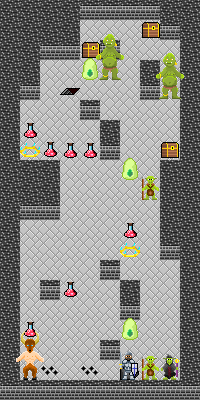}
        \caption{CAC-CF}
    \end{subfigure}
    \begin{subfigure}[t]{0.3\linewidth}
        \centering
        \includegraphics[width=.9\textwidth]{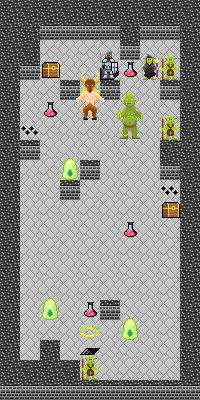}
        \caption{CAC-CAF}
    \end{subfigure}
    \begin{subfigure}[t]{0.3\linewidth}
        \centering
        \includegraphics[width=.9\textwidth]{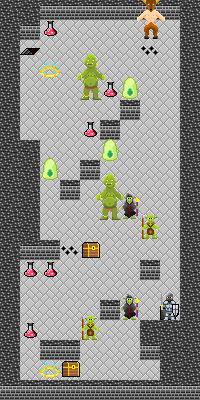}
        \caption{CAC-AF}
    \end{subfigure}

    \begin{subfigure}[t]{0.3\linewidth}
        \centering
        \includegraphics[width=.9\textwidth]{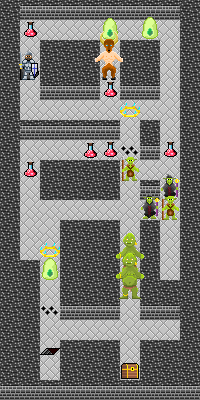}
        \caption{AC-CF}
    \end{subfigure}
    \begin{subfigure}[t]{0.3\linewidth}
        \centering
        \includegraphics[width=.9\textwidth]{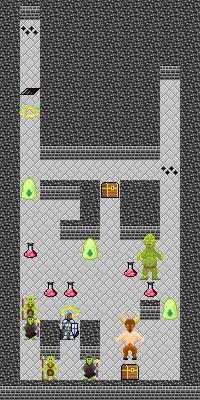}
        \caption{AC-CAF}
    \end{subfigure}
    \begin{subfigure}[t]{0.3\linewidth}
        \centering
        \includegraphics[width=.9\textwidth]{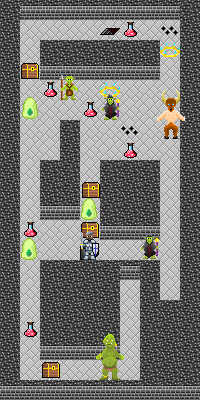}
        \caption{AC-AF}
    \end{subfigure}
    \caption{Generated MD2 levels for different furnishers (one per column) on the same layout per creator (one per row).}
    \label{fig:furnisher_samples}
\end{figure}

Figure~\ref{fig:furnisher_samples} shows three layouts created by the three creators, which are then furnished by each of the three furnishers. There are several consistent patterns among generator combinations: levels always have around 20 game elements (monsters comprise slightly less than half of those), potions are a bit more than half the number of monsters and a bit less than double the number of treasure chests. Differences in the layouts are also obvious, while the way in which each layout affects each furnisher less so. This section attempts to quantitatively analyze these patterns in creators, furnishers, and their combination.

\subsubsection{Layout Creators:}

Observing the layouts created by each creator, we decided to use the number of floor tiles, the length of the longest path between the player's starting position and the exit, and the number of wall chunks as metrics to analyze the generated layouts. The number of floor tiles just counts the number of empty tiles in the generated map, the longest path length calculates the number of moves that a player needs to traverse the map, and the number of wall chunks calculates the number of isolated wall segments in the generated layouts. 

\begin{figure}
    \centering
    \begin{subfigure}[t]{0.3\linewidth}
        \centering
        \includegraphics[width=\textwidth]{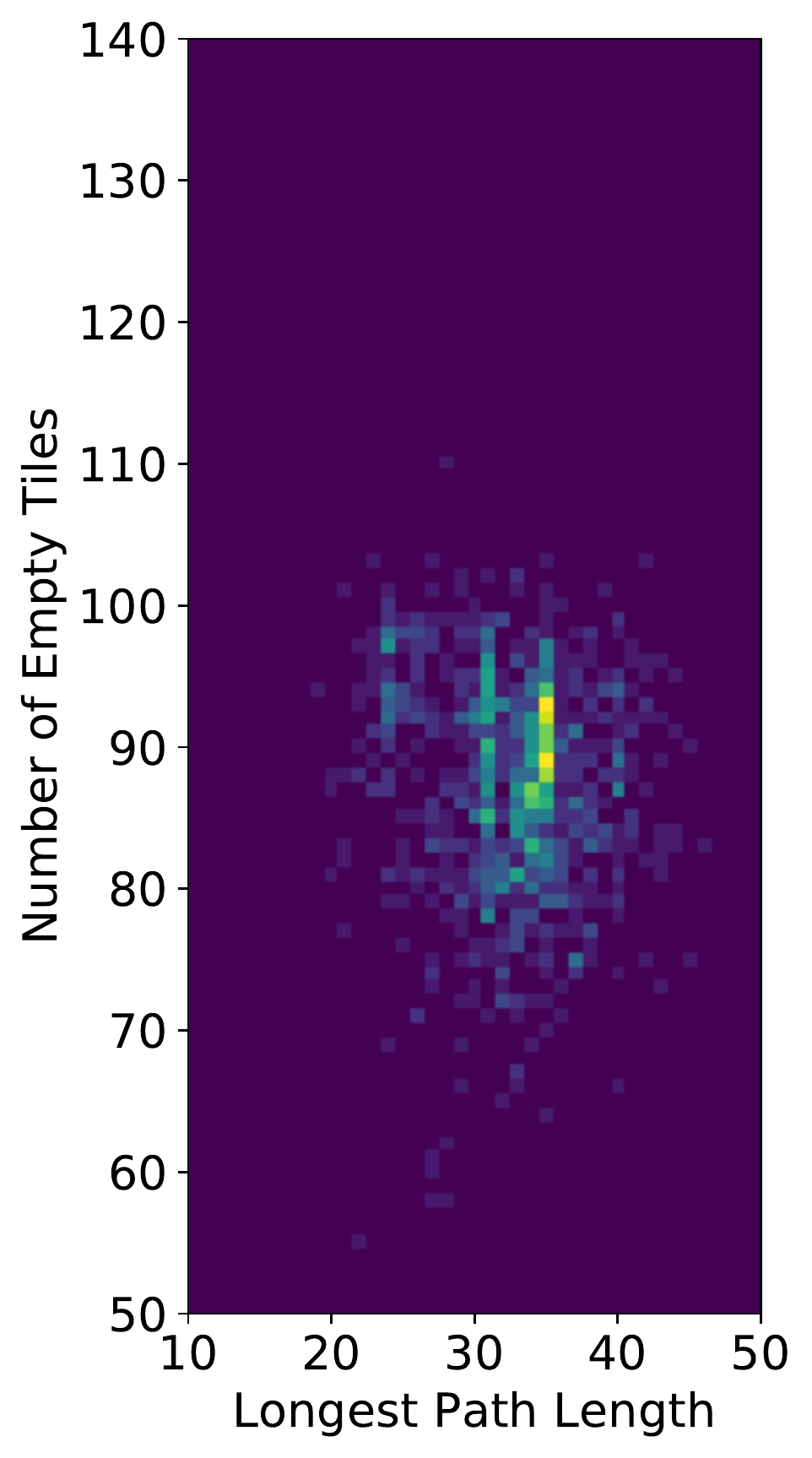}
    \end{subfigure}
    \begin{subfigure}[t]{0.3\linewidth}
        \centering
        \includegraphics[width=\textwidth]{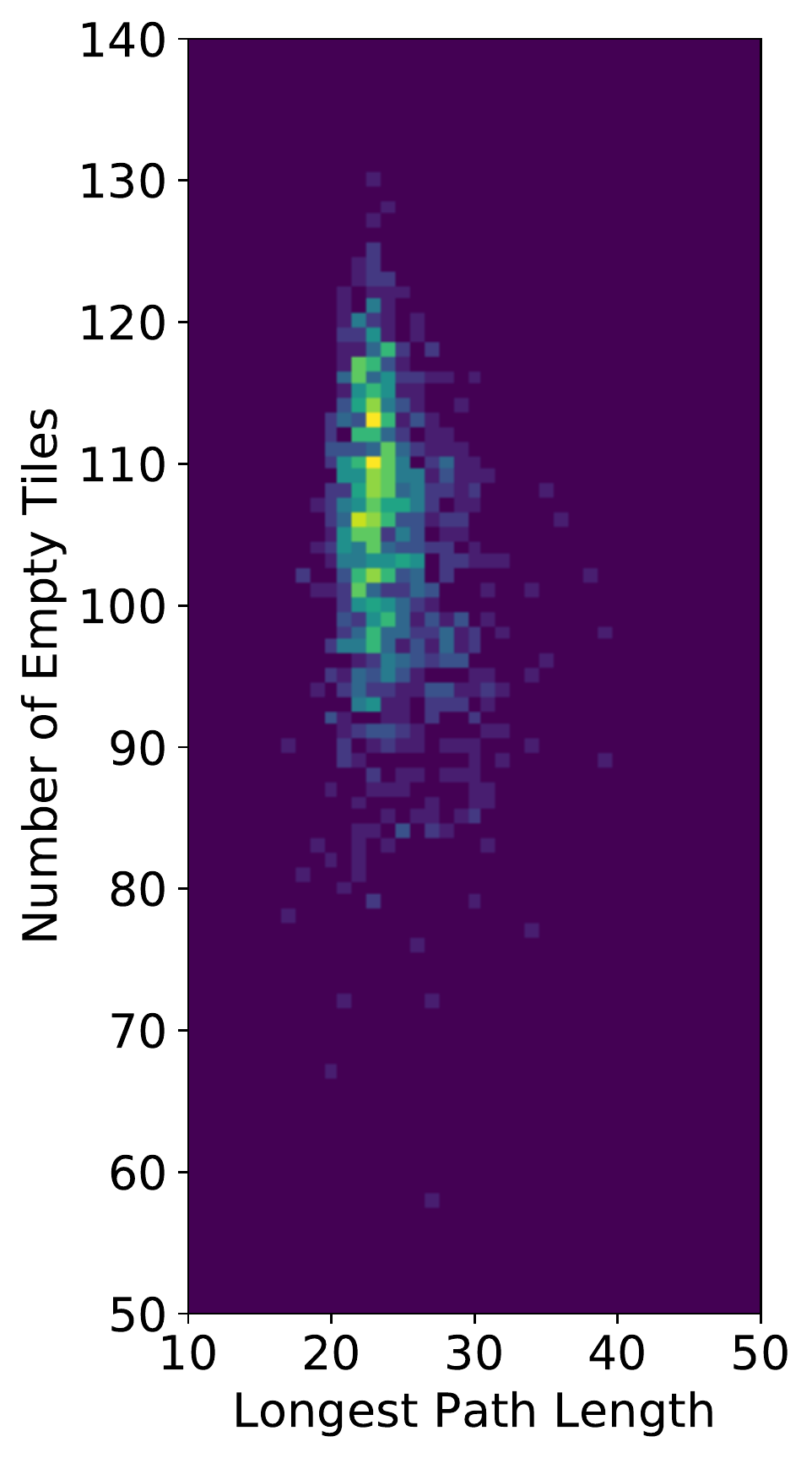}
    \end{subfigure}
    \begin{subfigure}[t]{0.3\linewidth}
        \centering
        \includegraphics[width=\textwidth]{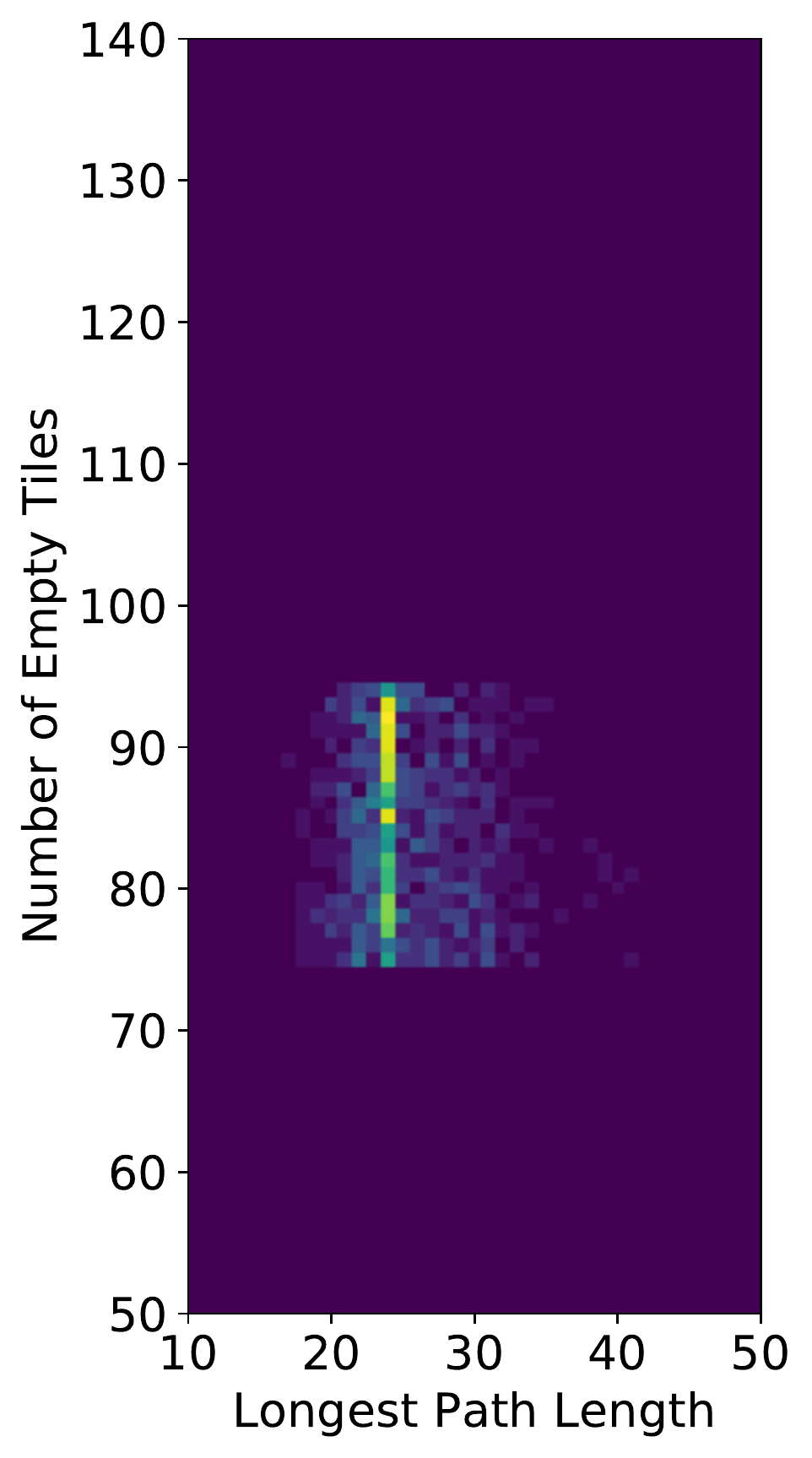}
    \end{subfigure}
    
    \begin{subfigure}[t]{0.3\linewidth}
        \centering
        \includegraphics[width=\textwidth]{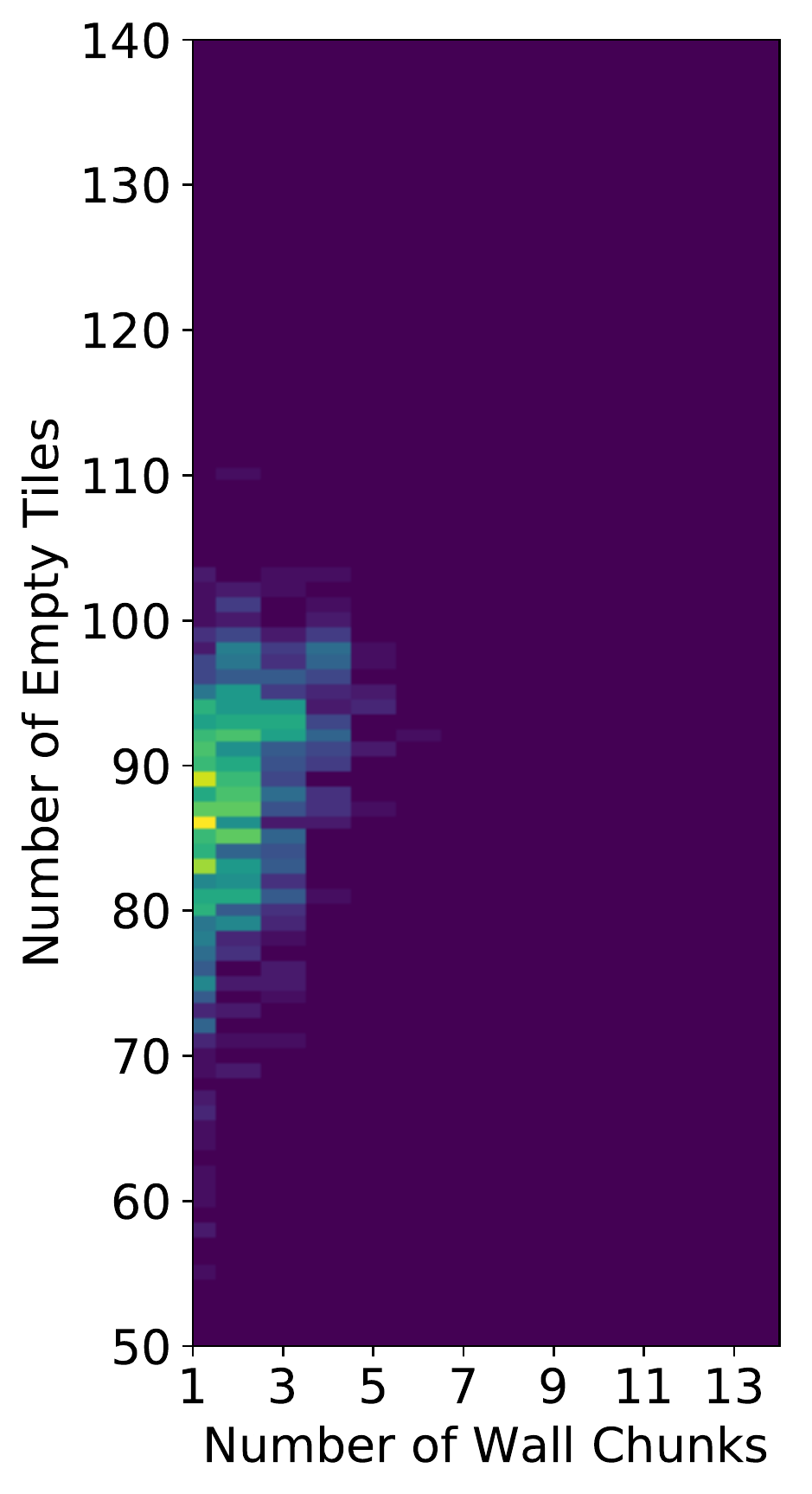}
    \end{subfigure}
    \begin{subfigure}[t]{0.3\linewidth}
        \centering
        \includegraphics[width=\textwidth]{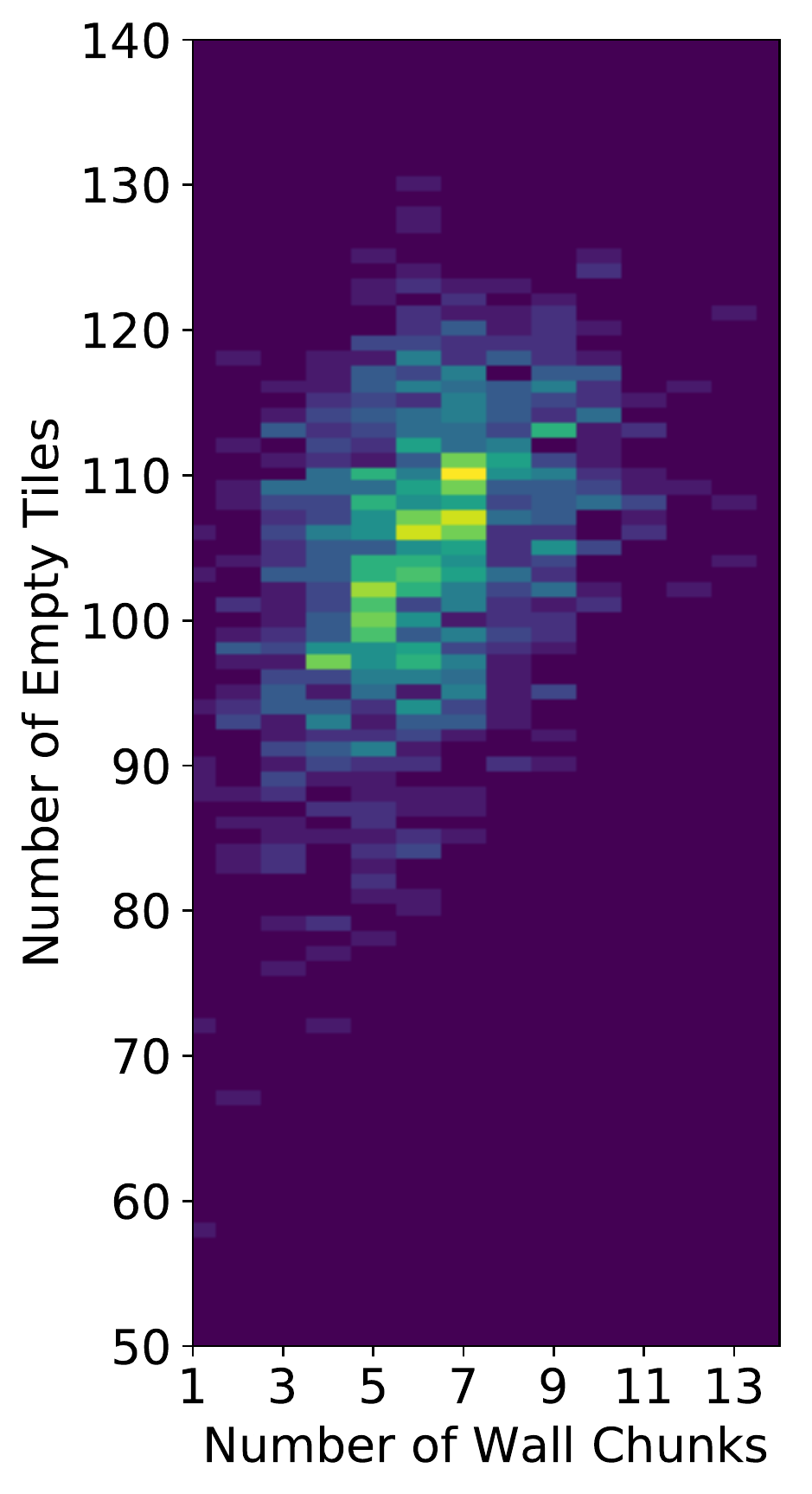}
    \end{subfigure}
    \begin{subfigure}[t]{0.3\linewidth}
        \centering
        \includegraphics[width=\textwidth]{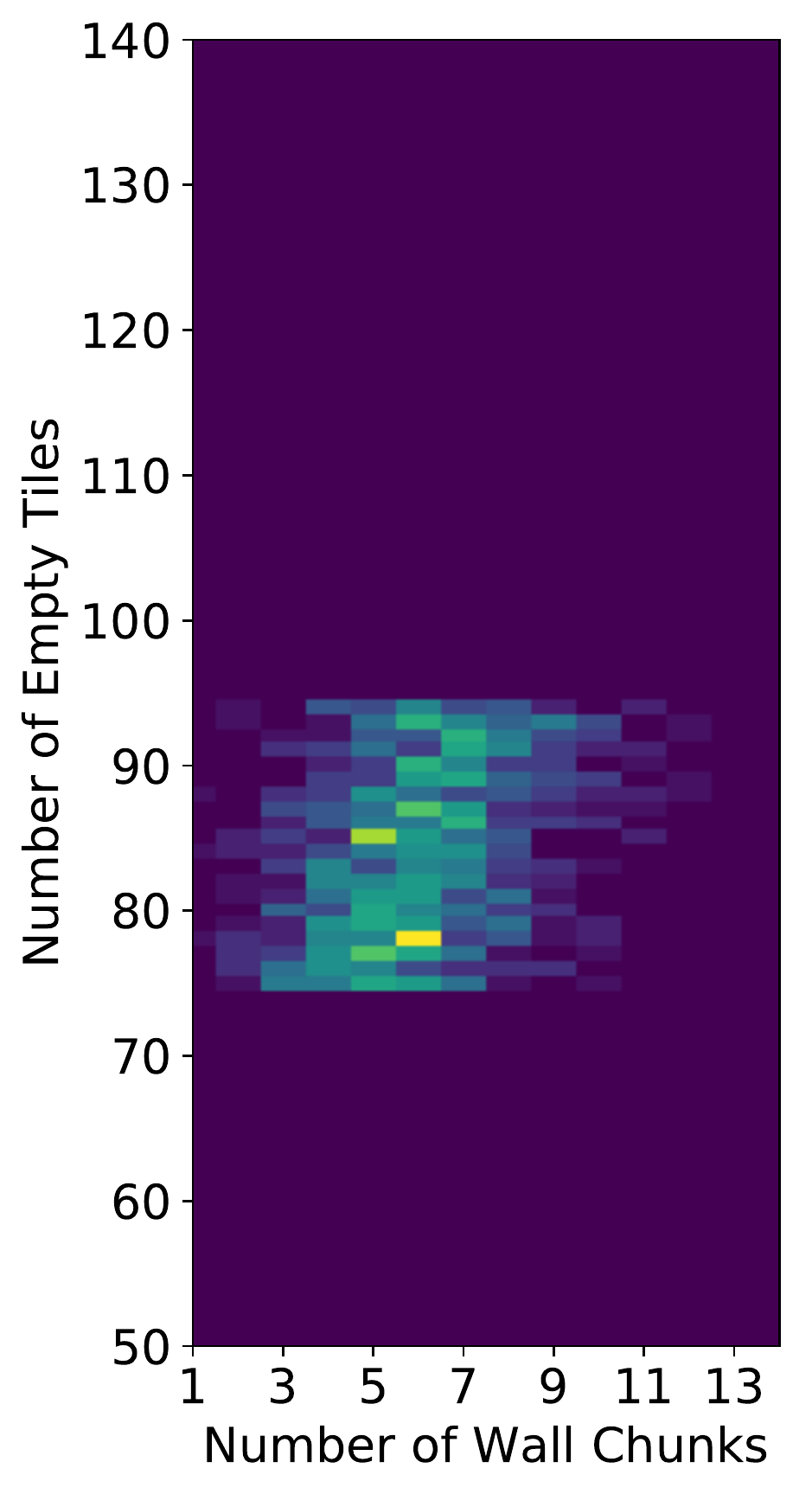}
    \end{subfigure}
    
    \begin{subfigure}[t]{0.3\linewidth}
        \centering
        \includegraphics[width=\textwidth]{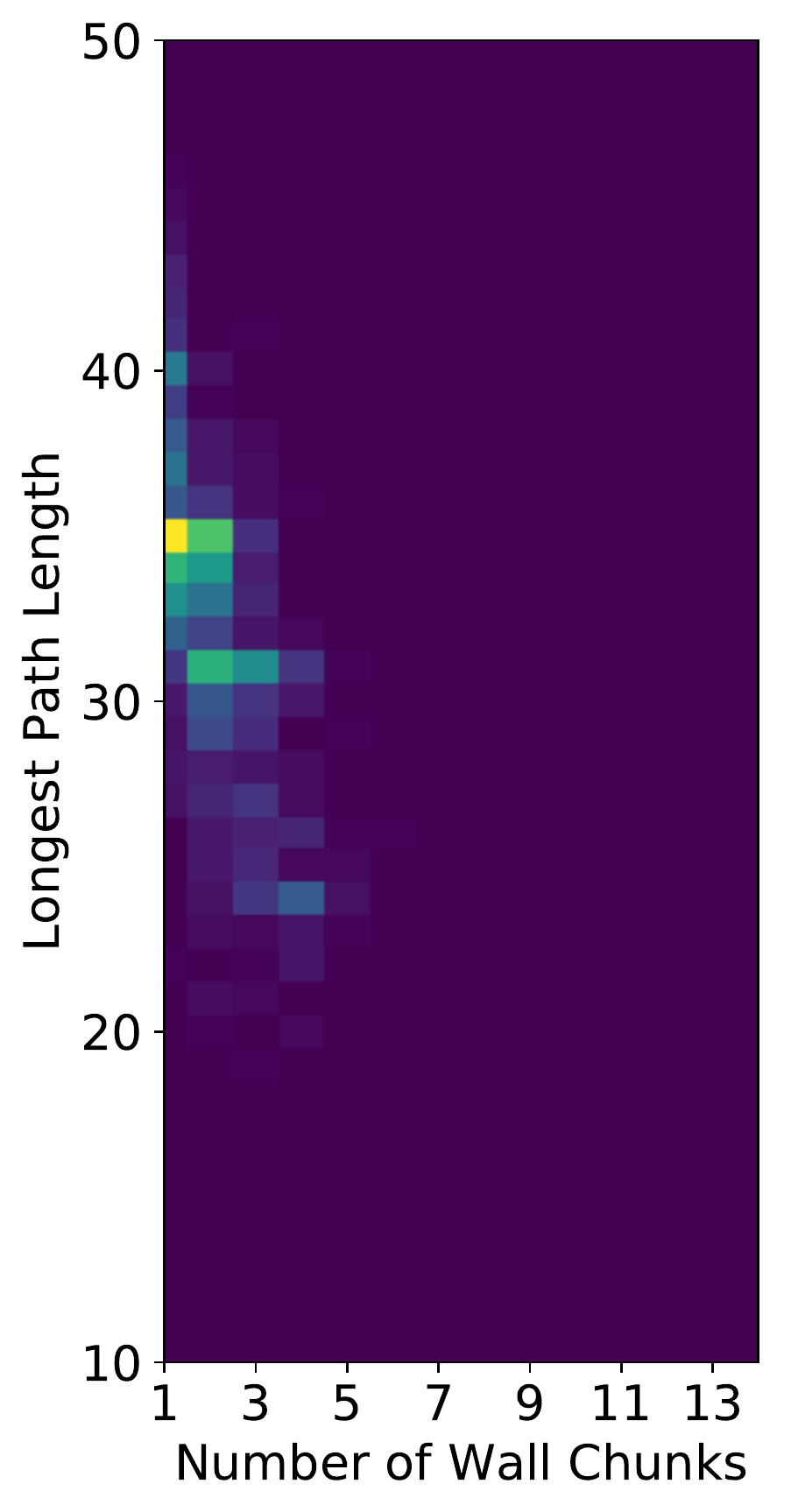}
        \caption{CC}
    \end{subfigure}
    \begin{subfigure}[t]{0.3\linewidth}
        \centering
        \includegraphics[width=\textwidth]{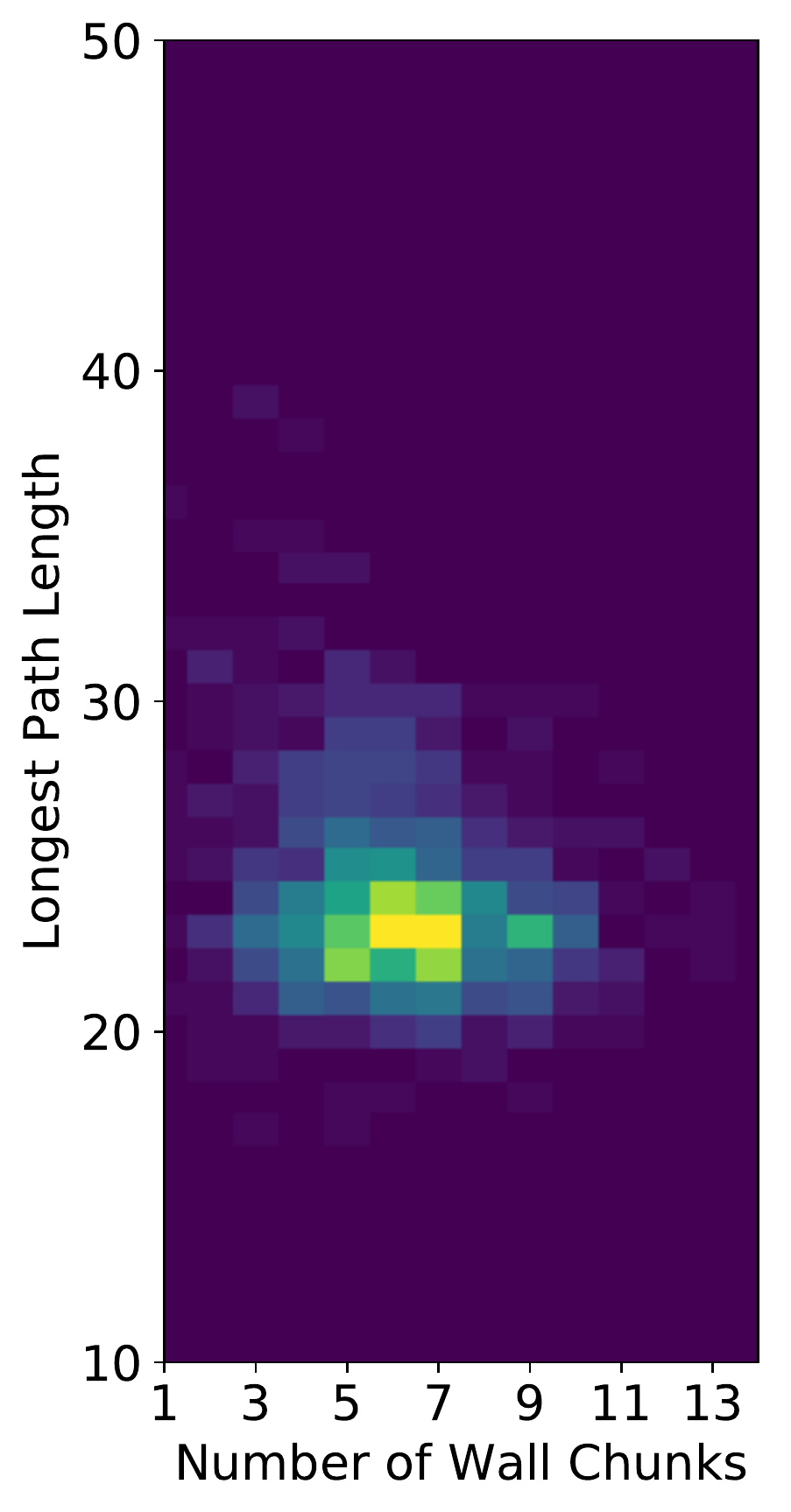}
        \caption{CAC}
    \end{subfigure}
    \begin{subfigure}[t]{0.3\linewidth}
        \centering
        \includegraphics[width=\textwidth]{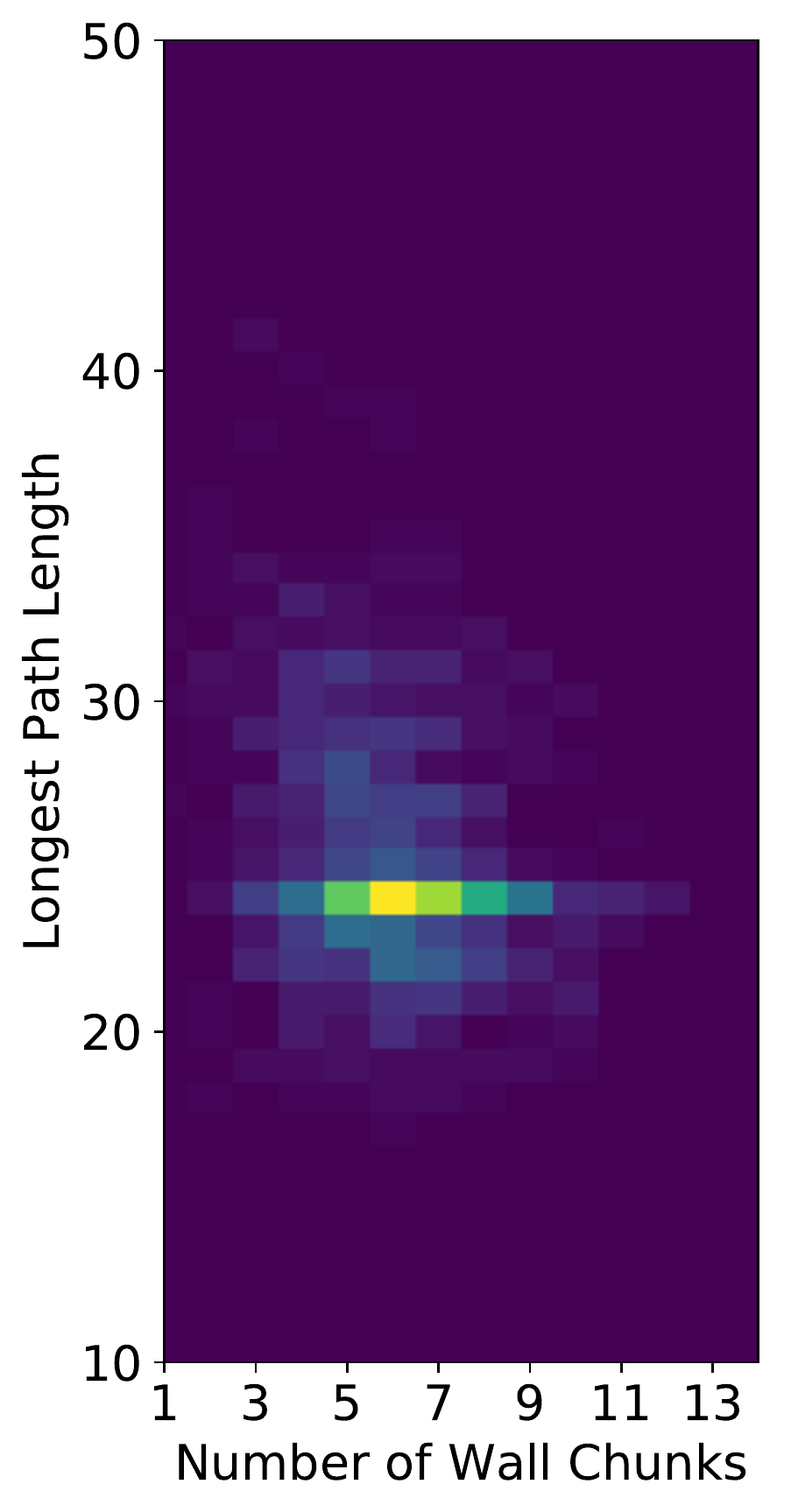}
        \caption{AC}
    \end{subfigure}
    \caption{The expressive range of the three different creators using the number of empty tiles, the longest path length, and the number of wall chunks.}
    \label{fig:creator_differences}
\end{figure}

Figure~\ref{fig:creator_differences} shows the expressive range of our three layout creators. It is obvious that the constrained-based creator generates levels with the longest path compared to the other two techniques. Also, it generates maps with the lowest number of isolated wall chunks. We think that the constraints satisfaction guarantees that rooms are not cutting each other, allowing for a longer path and fewer isolated wall chunks. On the other hand, CAC creates maps with the highest number of empty tiles which is due to the rules which govern the growth of cellular automata. Both AC and CAC generate maps with high amounts of isolated chunks, most likely due to the ability of the agent to cross its own path and the extra wall-seeding function in CA.

\subsubsection{Furnishers:}
The different combinations of creators and furnishers result in 9000 MD2 levels, out of which 3000 levels are generated by each creator (and different furnishers) and 3000 levels by each furnisher (and different creators). In order to assess how each algorithm affects the placement of game elements, these levels are grouped by creator or furnisher and their average metrics are compared with those of other creators or furnishers respectively. A broad range of metrics has been explored, including the number and ratio of all game elements, the path from entrance to exit and the number of potions or treasures which are guarded (i.e. their paths from the entrance are blocked by a monster). For the sake of brevity, only a subset of significant differences are reported here. 

\begin{figure}
    \centering
    \begin{subfigure}[t]{\linewidth}
        \centering
        \includegraphics[width=.9\textwidth]{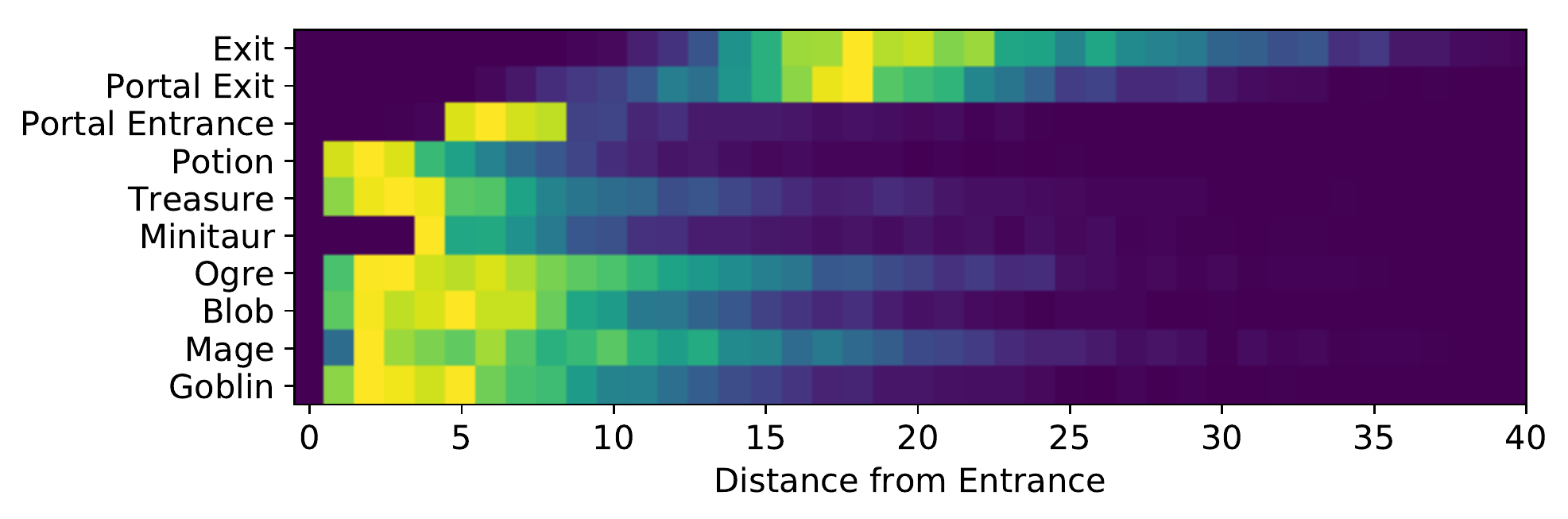}
        \caption{Constraint-based Furnisher}
        \label{fig:cf_Expressive}
    \end{subfigure}
    \begin{subfigure}[t]{\linewidth}
        \centering
        \includegraphics[width=.9\textwidth]{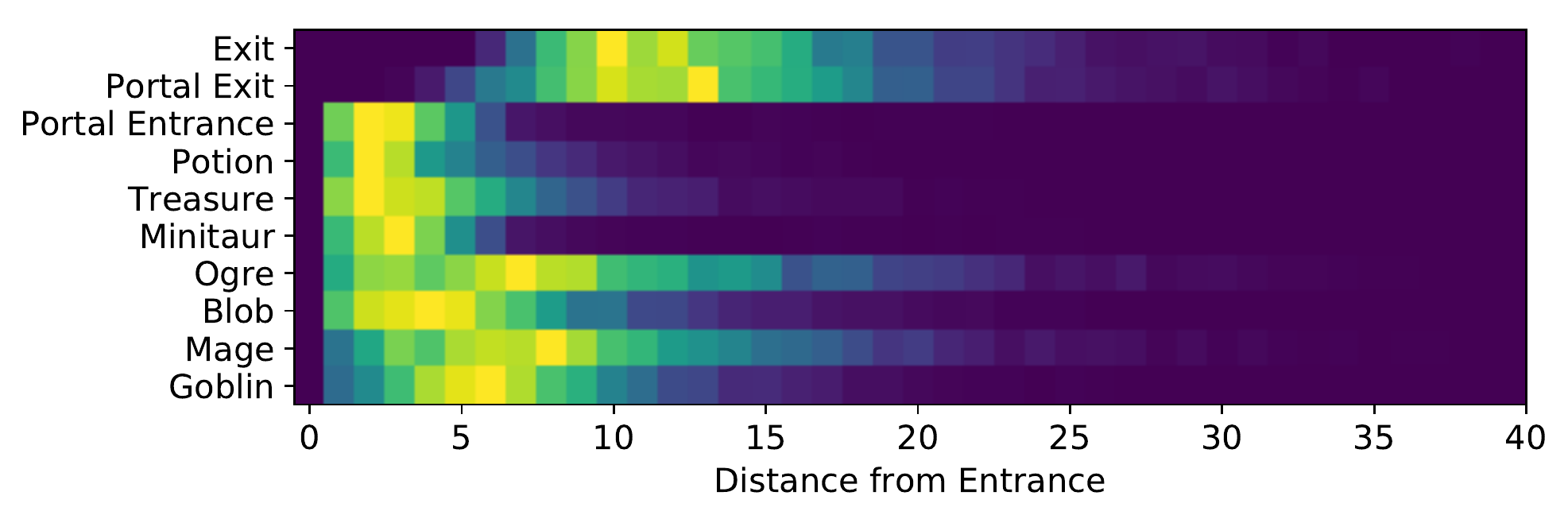}
        \caption{CA Furnisher}
        \label{fig:caf_expressive}
    \end{subfigure}
    \begin{subfigure}[t]{\linewidth}
        \centering
        \includegraphics[width=.9\textwidth]{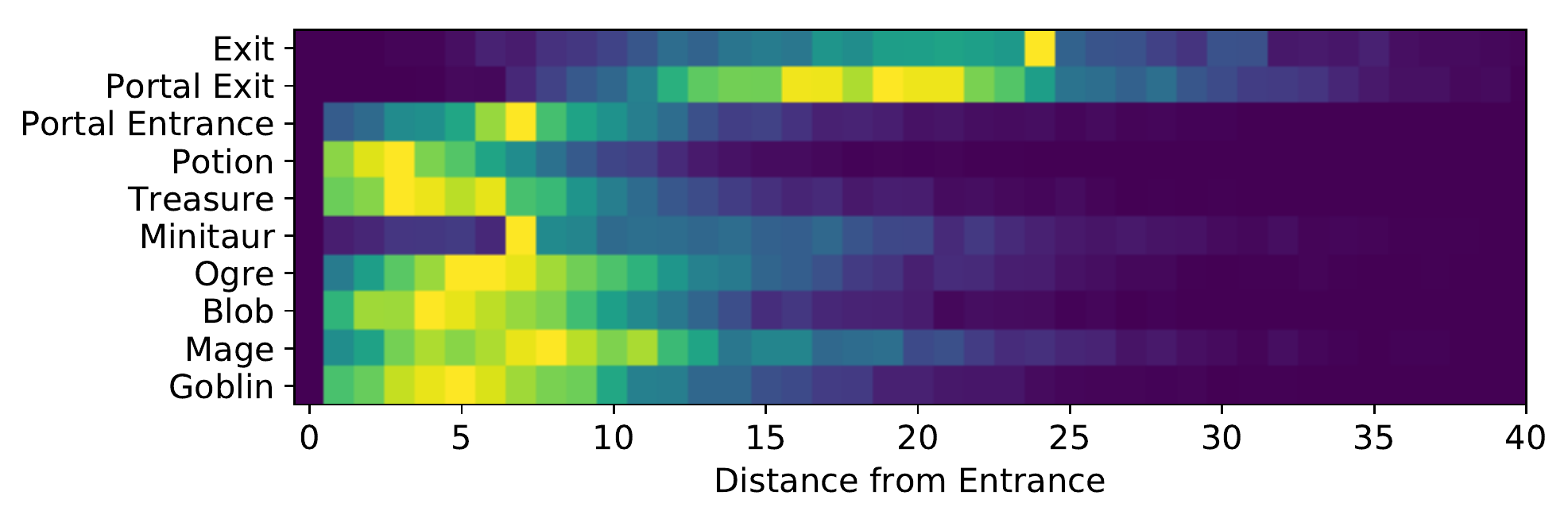}
        \caption{Agent-based Furnisher}
        \label{fig:af_expressive}
    \end{subfigure}
    \caption{Expressive range of the shortest distance from the entrance to all other game objects for each furnisher agent.}
    \label{fig:furnisherExpressive}
\end{figure}

Grouping MD2 levels by furnisher, Figure~\ref{fig:furnisherExpressive} shows the expressive range of these furnishers. We used the distance between the entrance and the various game objects as metrics for the analysis. The obvious difference that can be observed is that CAF has the smallest distance from entrance to exit or portal exit, compared to the two other furnishers. This is expected as CAF is the only furnisher which places both entrance and exit randomly  with the only control being on neighboring tiles, while other generators either greedily maximize the distance between entrance and exit (AF) or place them along the longest path (CF). Similarly, the agent-based furnisher has the largest entrance-minitaur distance compared to the other two techniques. In maps furnished by AF, the minitaur is most always able to maximize the distance between the entrance and itself.

\subsection{Playability metrics}
While structural differences shed light on the physical appearance of the levels, how these differences affect gameplay is an important next question. In order to gain some insight on how players would interact with generated levels, a number of artificial agents designed to represent different playstyles were used to playtest a sample of the 1000 levels generated for each creator-furnisher combination. These artificial agents, named procedural personas, use a variant of Monte-Carlo tree search~\cite{browne2012mcts} with an evolved exploration strategy described by~\cite{holmgard2018automated}. Three personas are used in this paper: the runner (which prioritizes reaching the exit with fewest actions), the monster killer (which prioritizes killing the most monsters and reaching the exit) and the treasure collector (which prioritizes collecting the most treasure and reaching the exit). To limit simulation times, results are calculated on simulations of procedural personas in the 100 first levels generated by each creator-furnisher combination.

\begin{figure}
    \centering
    \begin{subfigure}[t]{\linewidth}
        \centering
        \includegraphics[trim=0 20px 0 10px,width=\textwidth]{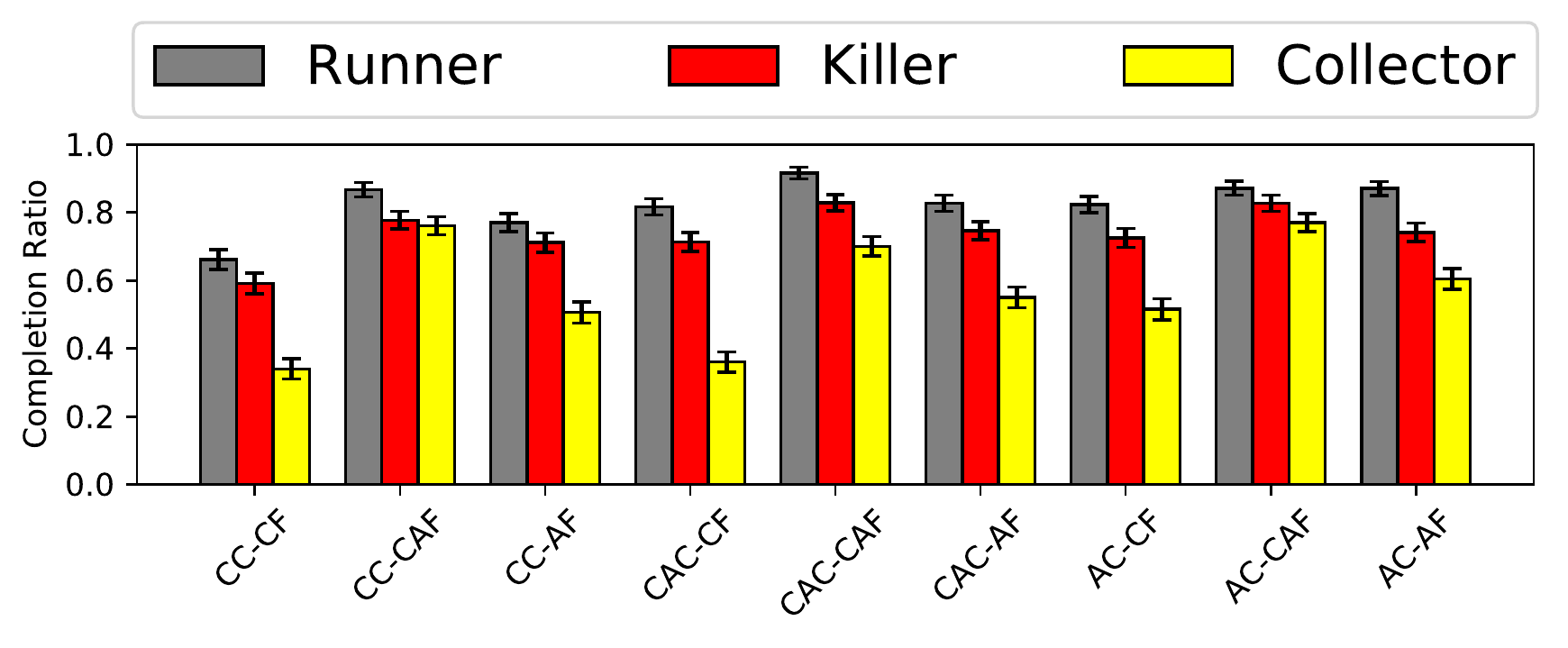}
        \label{fig:completionRate}
    \end{subfigure}
    \begin{subfigure}[t]{\linewidth}
        \centering
        \includegraphics[trim=0 20px 0 10px,width=\textwidth]{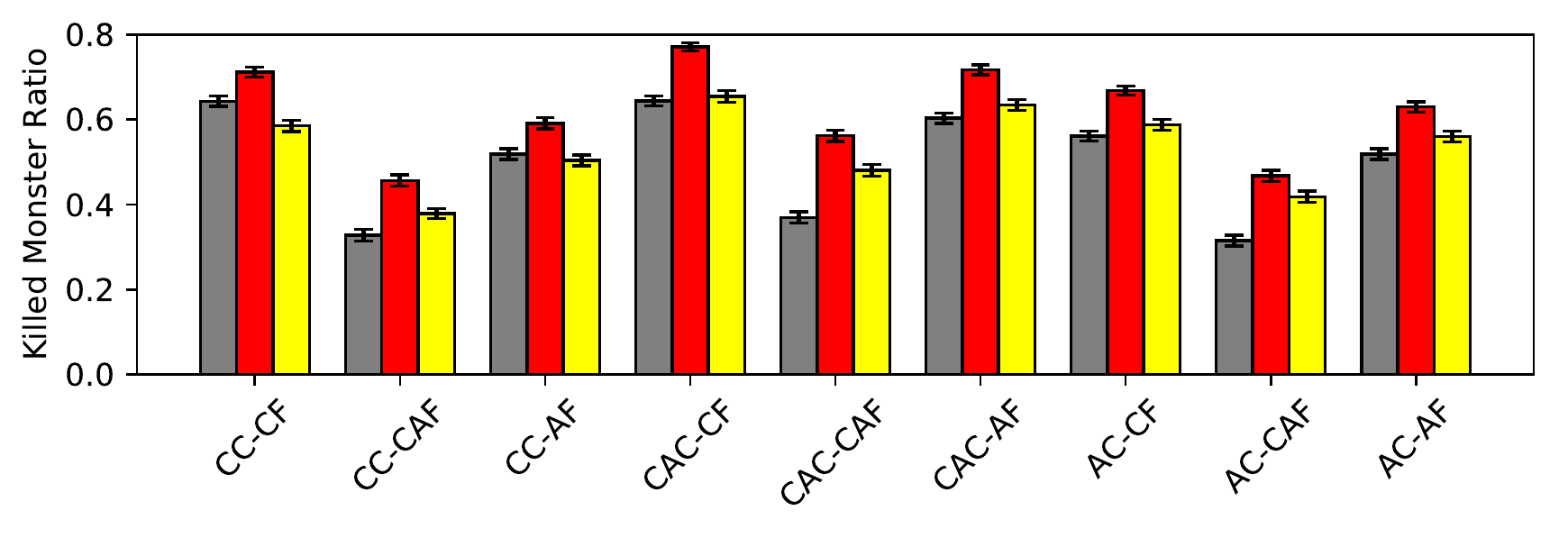}
        \label{fig:monstersKilled}
    \end{subfigure}
    \begin{subfigure}[t]{\linewidth}
        \centering
        \includegraphics[trim=0 20px 0 10px, width=\textwidth]{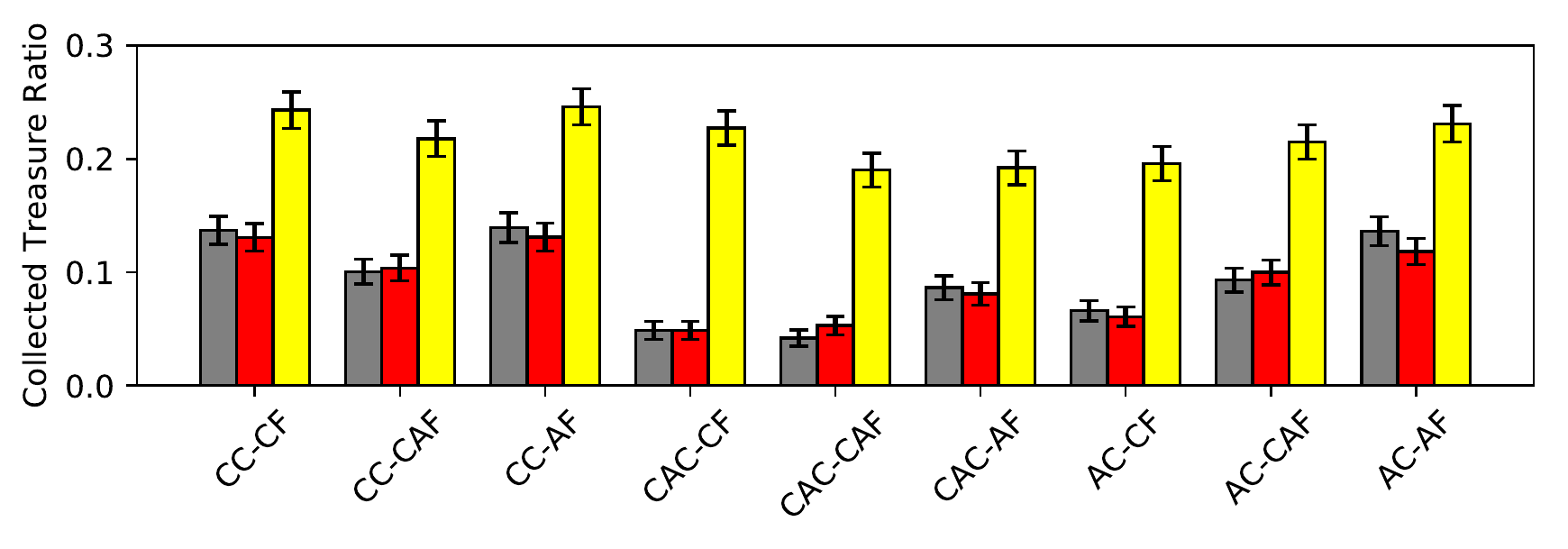}
        \label{fig:treasureCollected}
    \end{subfigure}
    \caption{Metrics of each persona on playthroughs of 100 levels; error bars show the 95\% confidence interval.}
    \label{fig:playthroughResults}
\end{figure}

Figure~\ref{fig:playthroughResults} summarizes three important metrics which are targeted explicitly by the personas (one each). An initial observation is that completion rates are high for runner and killer personas. Unsurprisingly, personas perform well on the metric they prioritize: runners reach the exit, monster killers (MK) kill more monsters, and treasure collectors (TC) collect more treasures. However, the differences between them are more prominent for some generators than for others: maps generated by CC-CF combination are the hardest to beat, and the biggest differences in treasures collected between the TC persona and the other two personas are found in  CAC-CF combination. 
It is also notable that certain creator-furnisher combinations result in generally more monsters killed (e.g. all combinations with CF), while levels by the cellular automata furnisher generally result in fewer kills. Similarly, CA creators have the least collected treasures for Runner and MK personas; on the other hand, the starker differences with the TC persona in this metric are with the constraint-based furnisher.

\section{Discussion}\label{sec:discussion}

Driven by the objective needs of the {MiniDungeons 2} game, which must produce new content while also operating on a mobile device, we designed and experimented with a multitude of computationally lightweight generators. Since the generators populate the level in steps, this allows for many combinations of patterns in the levels. As shown in the experimental analysis, the different generators introduce different patterns, and while visible differences in levels of Figure~\ref{fig:furnisher_samples} are mostly due to the layout creators, the number and placement of game objects seem less sensitive to differences among creators. Based on playthroughs of artificial agents, the generated levels allow for different strategies--- although some strategies do not guarantee that the level can be finished. The CC-CF combination created the hardest maps for any persona to beat, suggesting that they create more of a challenge than other combinations. Some generator types in particular penalize or reward specific playstyles (e.g. killing monsters in CAF levels) while some creators make a clearer distinction \textit{between} playstyles (treasure collection differences in CAC levels).

There are many directions for future work, but also on further analysis on how the patterns of each step of the generative pipeline affects patterns in the next. A more in-depth analysis using non-linear or possibly even computer vision techniques such as deep learning could shed more light on the dependencies of creator-furnisher pairings. Another idea would be to take a note from previous generative comparison work~\cite{cook2016danesh} to perform a more general comparison of metrics. Moreover, the artificial agents can be used as surrogate testers in order to fine-tune the placement rules of some objects, so that for instance the chance that runners can complete the generated levels increases while also increasing the number of treasures collected by treasure collectors (that currently rarely gain a third of all treasure in the level). We also believe that the Agent-based furnisher offers a fresh approach to the old problem of procedural map generation in games. In this paper, all agents move using simple one-step-look-ahead, but future work could install more rigorous tree search or localized optimization methods, allowing these agents to interact with one another in complex ways.

\section{Conclusion}\label{sec:conclusion}
This paper presented nine generative algorithms that decide either on the architecture or on the game object distribution of levels for {MiniDungeons2}. By combining these creators and furnishers, a broad set of patterns emerges. Experiments have demonstrated how levels created by different combinations of creators and furnishers differ from each other, and how they affect the playthroughs of artificial agents acting as surrogates of human players.

\section*{Acknowledgements}
Michael Cerny Green acknowledges the financial support of the SOE Fellowship from NYU Tandon School of Engineering. Ahmed Khalifa acknowledges the financial support from NSF grant (Award number 1717324 - "RI: Small: General Intelligence through Algorithm Invention and Selection.").

\bibliographystyle{ACM-Reference-Format}
\bibliography{constructive}
\end{document}